\documentclass[a4paper, 10pt, conference]{ieeeconf}      
\usepackage{FG2025}
\usepackage{graphicx}
\usepackage{amsmath}
\usepackage{calligra}
\usepackage{url}
\makeatletter
\let\NAT@parse\undefined
\makeatother
\usepackage{hyperref}

\FGfinalcopy 

\IEEEoverridecommandlockouts                              
\def\FGPaperID{162} 

\title{\LARGE \bf
GenEAva: Generating Cartoon Avatars with Fine-Grained Facial Expressions from Realistic Diffusion-based Faces
}

\author{\parbox{16cm}{\centering
    {\large Hao Yu$^1$, Rupayan Mallick$^2$, Margrit Betke$^1$, Sarah Adel Bargal$^2$}\\
    {\normalsize
    $^1$ Department of Computer Science, Boston University, Boston, USA\\
    $^2$ Department of Computer Science, Georgetown University, Washington, D.C., USA}
    }
}

\begin{document}

\ifFGfinal
\thispagestyle{empty}
\pagestyle{empty}
\else
\author{Anonymous FG2025 submission\\ Paper ID \FGPaperID \\}
\pagestyle{plain}
\fi
\maketitle

\begin{abstract}

Cartoon avatars have been widely used in various applications, including social media, online tutoring, and gaming. 
However, existing cartoon avatar datasets and generation methods struggle to present highly expressive avatars with fine-grained facial expressions and are often inspired from
real-world identities, raising privacy concerns. 
To address these challenges, we propose a novel framework, GenEAva, for generating high-quality cartoon avatars with fine-grained facial expressions. 
Our approach fine-tunes a state-of-the-art text-to-image diffusion model to synthesize highly detailed and expressive facial expressions. 
We then incorporate a stylization model that transforms these realistic faces into cartoon avatars while preserving both identity and expression.
Leveraging this framework, we introduce the first expressive cartoon avatar dataset, GenEAva 1.0, specifically designed to capture 135 fine-grained facial expressions, featuring 13,230 expressive cartoon avatars with a balanced distribution across genders, racial groups, and age ranges.
We demonstrate that our fine-tuned model generates more expressive faces than the state-of-the-art text-to-image diffusion model SDXL.
We also verify that the cartoon avatars generated by our framework do not include memorized identities from fine-tuning data.
The proposed framework and dataset provide a diverse and expressive benchmark for future research in cartoon avatar generation.
\end{abstract}

\section{INTRODUCTION}

Cartoon avatars have become increasingly important in various digital domains, serving as personalized digital representations in applications such as social media~\cite{nowak2018avatars}, chatbots~\cite{hadjiev2021evaluation}, online tutoring~\cite{fink2024ai,segaran2021does}, video conferencing~\cite{panda2022alltogether}, virtual reality~\cite{bimberg2024influence}, and video games~\cite{nizam2022avatar}. As digital communication evolves, cartoon avatars offer a compelling alternative to realistic human representations, providing users with enhanced personalization and privacy, and enriching user engagement and interaction across various platforms.

Despite the growing popularity of cartoon avatars across various applications, current avatar generation methods and cartoon face datasets have several limitations.
Many existing approaches struggle to create highly expressive cartoon avatars and fail to effectively convey nuanced emotions~\cite{shen2023overview}. This is partly due to the lack of cartoon face data with diverse facial expressions, as most available datasets predominantly feature neutral or basic facial expressions~\cite{cartoonset}. Furthermore, generative models sometimes memorize identities or generate avatars that resemble real individuals from the training data rather than producing genuinely novel identities~\cite{277172}, which raises significant privacy concerns. An ideal avatar generation system should create diverse and unique representations without overfitting to specific individuals. Additionally, cartoon face datasets often exhibit bias in terms of age and race, with many skewed toward young, lighter-skinned characters, \textit{e.g.}, Manga109~\cite{fujimoto2016manga109}. 

In light of these challenges, we propose a novel framework, GenEAva, for \textbf{Gen}erating \textbf{E}xpressive cartoon \textbf{Ava}tars. 
In this work, we specifically define cartoon avatars as cartoon-style digital representations that represent identities in the images generated by the diffusion model.
We first propose a facial expression generation model that can generate fine-grained realistic facial expressions across 135 emotion categories based on fine-tuning a text-to-image (T2I) diffusion model. 
High-quality realistic facial expressions are then generated using this model with carefully designed text prompts. The prompts are curated using the state-of-the-art Large Language Model (LLM) GPT-4o~\cite{hurst2024gpt}, ensuring a wide range of age groups, equal representation of males and females, and a balanced racial distribution across seven racial groups. 
Finally, we convert these realistic facial expressions into cartoon avatars through a stylization method while maintaining the identity and facial expression of the original faces. Additionally, we present a comprehensive evaluation pipeline for cartoon avatar generation, focusing on facial expression fidelity and representation, identity memorization, and the preservation of identity and expression during stylization.

We present GenEAva 1.0, the first cartoon avatar dataset that is specifically designed to include fine-grained facial expressions with unique identities, diverse age groups, and a balanced racial distribution.
GenEAva 1.0 consists of 13,230 cartoon avatars of 135 facial expressions.
We conducted extensive experiments to evaluate the quality of the cartoon avatars in the dataset and show that the generated images in GenEAva 1.0 present fine-grained facial expressions, surpassing the state-of-the-art T2I diffusion model SDXL~\cite{podell2023sdxl} across various visual quality metrics.
We also demonstrate that the dataset includes novel identities without instances of memorization from the fine-tuning dataset through quantitative analysis and a user study.
Finally, we validate that fine-grained facial expressions and novel identities are maintained through the stylization module using both quantitative analysis and a user study. 

We summarize our contributions as follows:
\begin{itemize}
\item 
We propose a novel framework, GenEAva, for the generation of expressive cartoon avatars from realistic faces generated by T2I diffusion models.
\item
We fine-tune the state-of-the-art T2I SDXL diffusion model to generate 
particularly fine-grained facial expressions.
\item 
We propose a diverse cartoon avatar dataset, GenEAva 1.0, with fine-grained facial expressions, unique identities, and balanced age, gender, and racial distribution. 
\end{itemize}

\section{RELATED WORK}

Our work lies at the intersection of facial expression generation, memorization in generative models, and synthetic face data. 
We review the related literature for each domain.

\subsection{Facial Expression Generation}

Facial expression generation refers to the process of synthesizing or modifying facial expressions in images or videos. Earlier approaches for facial expression generation are based primarily on Generative Adversarial Networks (GANs)~\cite{goodfellow2020generative}. For example, StarGAN~\cite{choi2018stargan} enables facial image editing with basic expressions through multi-domain image-to-image translation. To achieve more fine-grained control over facial expressions, several methods have incorporated Action Units (AUs) for more precise expression manipulation~\cite{pumarola2018ganimation, tripathy2020icface, zhao2021action}. GANmut~\cite{d2021ganmut} further learns an expressive and interpretable conditional space of emotions to generate compound emotions. EmoStyle~\cite{azari2024emostyle} uses StyleGAN2~\cite{karras2020analyzing} with a valence-arousal space for intuitive, continuous expression control.

More recently, diffusion models~\cite{ho2020denoising} have achieved significant success in generating high-quality images and have been applied to facial expression generation. Stable Diffusion~\cite{rombach2022high} has shown an exceptional ability to generate high-quality images from text prompts with basic facial expressions. Pikoulis \textit{et al.}~\cite{pikoulis2023photorealistic} fine-tuned Stable Diffusion using CLIP latent guidance to generate seven basic emotions. The most relevant work to ours is by Liu \textit{et al.}~\cite{liu2024towards}, which queries a dataset of 135 expressions and transfers the specific facial expression using a conditional diffusion model. In contrast, our approach does not rely on selecting a reference image for expression transfer. Instead, we fine-tune the diffusion model to learn the distribution of 135 facial expressions and sample directly from it, which enables the generation of novel images with intended facial expressions.

\subsection{Identity Memorization in Generative Models}

Memorization in a generative model refers to the generation or reproduction of the training data by a trained model at the time of inference. Initial empirical studies involving GANs \cite{Webster2021ThisP} have challenged the novelty of generation using a kind of membership inference attack. Later, generative models such as diffusion models have also been shown to replicate the underlying training priors at the time of generation. Somepalli \textit{et al. }\cite{somepalli2022diffusion} identified direct replication of training data by stable diffusion. Furthermore, Carlini \textit{et al.} \cite{carlini_usenix} demonstrated the lack of privacy preservation in the diffusion models, showing the leak in the training data at the inference time. Carlini further showed that replication can be induced by prompting the image captions from the LAION dataset, which is used for training diffusion models. Several works studied the mitigation of memorized models \cite{chen_memorisation}, \cite{somepalli2023understanding}
to reduce the regeneration of training data. Other methods include differentially private generative models for privacy-preserving image generation methods \cite{bie2023private}, \cite{chen_privacy}, \cite{dockhorn2023differentially}. 
In this work, we focus on identity memorization, which is a specific type of memorization in generative models where the generated face images replicate real identities from the training data.


\subsection{Synthetic Face Data}

Recent advances in generative models (\textit{e.g.}, Generative Adversarial Networks~\cite{goodfellow2020generative,sauer2022stylegan}, Diffusion Models~\cite{ho2020denoising,rombach2022high}) have enabled the creation and use of synthetic face data for face-related computer-vision tasks such as face recognition, landmark localization, and face parsing~\cite{bae2023digiface1m,Sface_Boutros,kim2023dcface,melzi2023gandiffface,qiu2021synface,wood2021fake}. 
Synthetic face datasets like DigiFace-1M~\cite{bae2023digiface1m}, the Face Synthetics Dataset~\cite{wood2021fake}, DCFace Synthetic Dataset~\cite{kim2023dcface}, and the GANDiffFace Dataset~\cite{melzi2023gandiffface} provide millions of synthetic faces with diverse attributes, which serve as alternatives to real-world datasets, mitigating privacy and bias issues. 
However, these synthetic face datasets are mostly designed for face detection and recognition, and include only limited facial expressions.

Studies have also explored the use of synthetic data for facial expression recognition~\cite{abbasnejad2017using,mishra2023enhancing,he2024synfer}. For example, Abbasnejad \textit{et al.}~\cite{abbasnejad2017using} used a 3D face model to generate six basic facial expressions. SynFER~\cite{he2024synfer} creates synthetic facial expression data with generative models based on facial action units and text guidance. Although these studies limit themselves to six basic facial expressions, we present a pipeline and dataset of 135 fine-grained facial expressions.


Several datasets have been specifically developed for cartoon faces.
IIIT-CFW~\cite{mishra2016iiit} contains 8,928 cartoon faces of 100 public figures. WebCaricature~\cite{huoa2017webcaricature} includes 6,042 caricatures and 5,974 photographs annotated with 17 facial landmarks. Manga109~\cite{fujimoto2016manga109} is a collection of 109 Japanese comic books designed for face detection tasks, and Danbooru~\cite{branwen2019danbooru2019} comprises over 970k anime images from 70k identities. Cartoon Set~\cite{cartoonset} provides sets of 10k and 100k 2D cartoon avatar images. The iCartoonFace Dataset~\cite{zheng2020cartoon} includes 389,678 images of 5,013 cartoon characters annotated for face detection and recognition. 
To the best of our knowledge, the cartoon face dataset that we introduce is the first dataset
that specifically focuses on presenting cartoon face images with fine-grained facial expressions across 135 categories.

\section{METHOD}

\begin{figure*}[t]
    \centering
    \includegraphics[width=\textwidth]{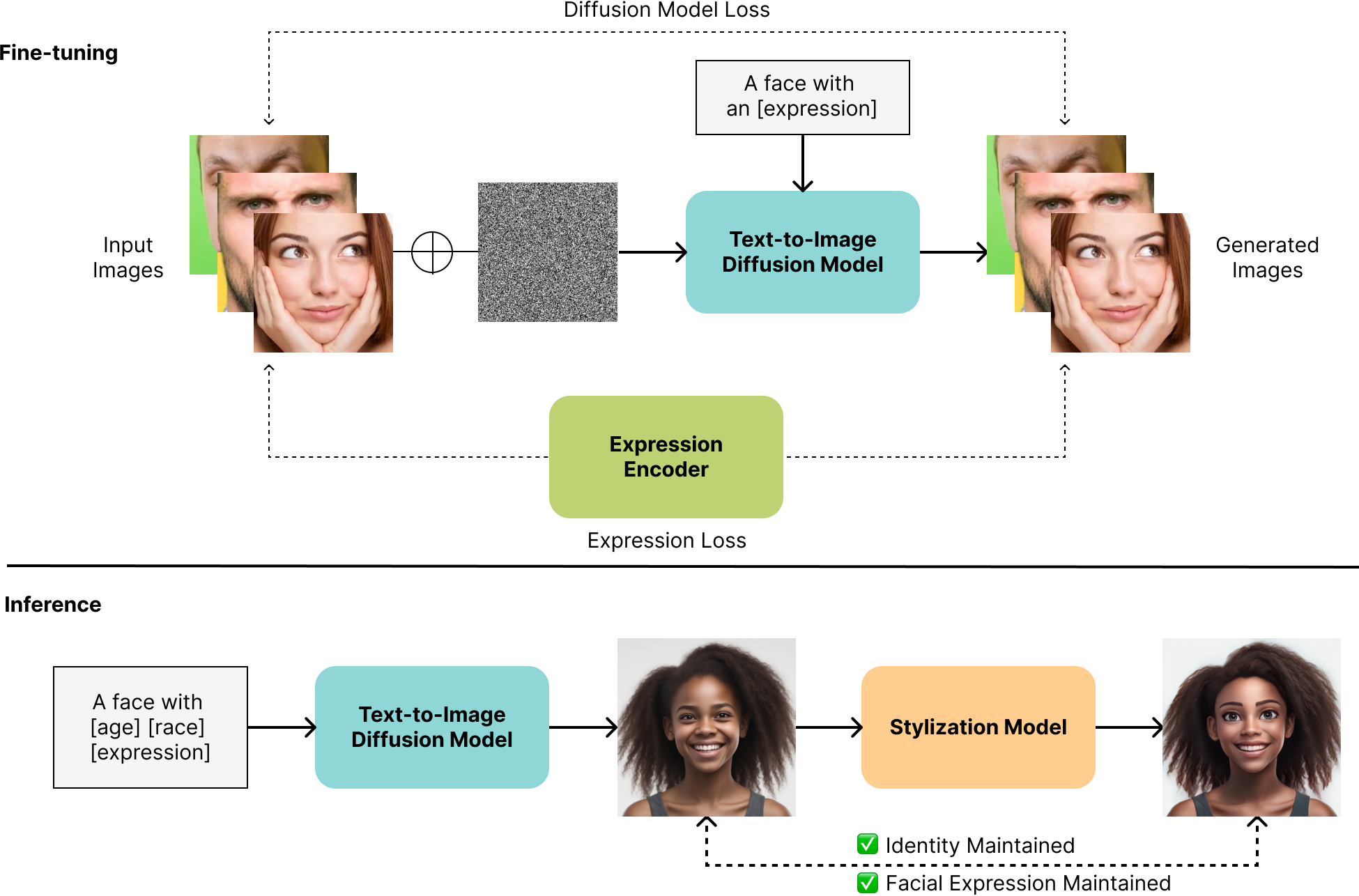}
    \caption{The proposed Pipeline, GenEAva, for generating expressive cartoon avatars. During the fine-tuning phase, we train a text-to-image diffusion model using facial expression images. The model is optimized with a combination of diffusion model loss (DM loss) and expression loss computed by an expression encoder. In the inference phase, we generate facial expression images by prompting the model, followed by applying a stylization model to transform them into cartoon avatars. 
    }
    \label{fig:pipeline}
\end{figure*}

We propose GenEAva, a novel pipeline for generating high-quality cartoon faces with fine-grained facial expressions. As shown in Figure~\ref{fig:pipeline}, the pipeline includes two phases: fine-tuning and inference. We first build a facial expression generation model that can generate photorealistic faces with fine-grained facial expressions based on a state-of-the-art text-to-image (T2I) diffusion model. We finetune a pretrained T2I model on 135 classes of facial expressions using both a diffusion model loss and an expression loss.
At inference, we generate high-quality face images with diverse facial expressions by prompting the T2I model and then utilize a stylization model to convert these photorealistic faces into cartoon avatars. 

\subsection{Preliminary on Text-to-Image Diffusion Models} 
Text-to-Image (T2I) diffusion models represent an emerging class of generative models that have recently achieved impressive results on various generative modeling benchmarks~\cite{saharia2022photorealistic,zhang2023adding}. They consist of a forward process and a reverse process~\cite{ho2020denoising}. In the forward process, a data sample~$\mathbf{x}_0$ is incrementally converted into pure Gaussian noise over a series of diffusion steps, where Gaussian noise $\epsilon \sim \mathcal{N}(0, 1)$ is gradually added at each step $t$, resulting in a sequence of intermediate noisy images $\mathbf{x}_t$. The reverse diffusion process begins with a standard Gaussian distribution and iteratively removes noise to generate a sample that resembles the training distribution. For training efficiency, these processes are always operated in a latent space $\mathbf{z} = \mathcal E(\mathbf{x})$ obtained using an image encoder $\mathcal E$. Given a timestep $t$, an intermediate noisy latent feature $\mathbf{z}_t$, and a text feature $\tau(p)$, generated by a text encoder $\tau$ and a text prompt $p$, a T2I diffusion model trains a conditional denoising U-Net~\cite{ronneberger2015u} to predict the added noise~$\epsilon$ using the squared error loss 
\begin{equation}
\begin{aligned}
    \mathcal{L}_{\text{dm}} &= \|\epsilon - \epsilon_\theta(\mathbf{z}_t, t, \tau(p))\|^2_2. \label{eq:dmloss}
\end{aligned}
\end{equation}
The closed form of $\mathbf{z}_t$ can be derived as:
\begin{equation}
    \begin{aligned}
    \mathbf{z}_t &= \sqrt{\bar{\alpha}_t} \mathbf{z}_0 + (1 - \bar{\alpha}_t) \epsilon, \\
    \mathrm{with} \:\:\bar{\alpha}_t &= \prod_{s=1}^{t} \alpha_s \: \: \: \mathrm{and} \quad \epsilon \sim \mathcal{N}(0, 1),\label{eq:zt}
\end{aligned}
\end{equation}
where $\alpha_s$ is a sequence of predefined coefficients controlling the variance of noise added at each step. 
Trained on large-scale image-text datasets such as LAION-5b~\cite{schuhmann2022laion}, T2I diffusion models are capable of generating high-quality images from text descriptions. 

\subsection{Expression-guided Finetuning}

To enable the T2I diffusion model to generate accurate, diverse, and fine-grained facial expressions, we finetune a state-of-the-art pre-trained T2I diffusion model, SDXL~\cite{podell2023sdxl}, on in-the-wild facial expression images. 

\paragraph{Finetuning on Diverse Expression Images}
We adopt Emo135~\cite{chen2022semantic}, a face dataset containing 135 fine-grained expression categories. For the best generation quality, we perform a series of pre-processing steps. First, we detect and crop faces in the images using the face detector RetinaFace~\cite{deng2019retinaface}. Since some images in the dataset contain watermarks, we use a watermark detection algorithm~\cite{watermark-detection} to filter out such images. Additionally, to address the imbalance of facial expressions in the original dataset, we construct a balanced dataset of 135 facial expressions by randomly sampling from the original dataset. This process results in a curated dataset containing 1,080 images, which is then used to finetune the SDXL model.

\paragraph{Training objectives}
While standard diffusion models are trained using a squared loss as in Eq.~\ref{eq:dmloss}, we incorporate an expression loss $\mathcal{L}_{\text{exp}}$ to guide the model toward generating more accurate and nuanced facial expressions. We use an expression encoder $\mathcal{E}_{\text{exp}}$ to extract the expression representation of an image. Specifically, we employ a state-of-the-art facial expression recognition model, POSTER~\cite{zheng2023poster}, as our expression encoder.
To guide the generation process, we compute the mean squared error (MSE) between the expression representations of the generated image $\hat{\mathbf{x}}_0$ and the real image $\mathbf{x}_0$. The expression loss is defined as:
\begin{equation}
    \mathcal{L}_{\text{exp}} = \text{MSE}(\mathcal{E}_{\text{exp}}(\mathbf{x}_0), \mathcal{E}_{\text{exp}}(\hat{\mathbf{x}}_0)).
\end{equation}
As derived previously~\cite{ho2020denoising}, we can approximate $\hat{\mathbf{x}}_0$ at any timestep~$t$ by a one-step reverse formula from Eq.~\ref{eq:zt}, which is defined as:
\begin{equation}
    \begin{aligned}
     \hat{\mathbf{z}}_0 = \frac{\mathbf{z}_t - \sqrt{1 - \bar{\alpha}_t} \epsilon_\theta}{\sqrt{\bar{\alpha}_t}}, \\
     \hat{\mathbf{x}}_0 = \mathcal{D}(\hat{\mathbf{z}}_0),
    \end{aligned}
\end{equation}
where $\mathcal{D}$ is an image decoder. 
The overall training objective is a combination of the original diffusion model loss and the expression loss with an $\alpha$ scaling factor, formulated as:
\begin{equation}
    \mathcal{L} = \mathcal{L}_{\text{dm}} + \alpha \mathcal{L}_{\text{exp}}. 
\end{equation}

\subsection{Facial Expression Generation}

Given the fine-tuned SDXL model, we generated high-quality facial expression images through carefully designed prompts. These prompts were specifically tailored to ensure a balanced representation of gender, age, and race in the generated images. We leveraged the state-of-the-art large language model GPT-4o~\cite{hurst2024gpt} to generate prompts that include equal representation of males and females, cover a wide range of ages from teenagers to elderly individuals, and provide a balanced representation across seven racial groups (White, Black, Indian, East Asian, Southeast Asian, Middle East, and Latino~\cite{karkkainen2021fairface}).
A sample prompt is: \textit{``A photorealistic face of a middle-aged Indian woman with shoulders visible, displaying a facial expression of delight, plain white background."}

To ensure consistent image quality, 
we filtered out images where the face appears too close to the camera, resulting in incomplete facial features, as well as images containing multiple generated faces.

\subsection{Cartoon Style Transfer}

Lastly, a face stylization method is applied to convert the generated images into a cartoon style. 
The framework allows for the integration of alternative stylization methods, and the evaluation pipeline presented in Section~\ref{sec:experiments} can still be applied for assessing the entire framework.
In this study, we use DCTNet~\cite{men2022dct}, a state-of-the-art image translation architecture for few-shot portrait stylization. We selected it for its advanced ability to synthesize high-fidelity content and its strong generality. It was trained to synthesize artistic portraits in various styles. We employed its pre-trained model for 3D cartoon style to generate cartoon avatars with fine-grained facial expressions. 
We then validated that fine-grained facial expressions are maintained through the stylization module using a user study presented in Section~\ref{sec:style_userstudy}.

\begin{figure*}[t]
    \centering
    \rotatebox{90}{\hspace{0.7cm}Cartoon\hspace{1cm} Realistic}
    \includegraphics[width=0.95\textwidth]{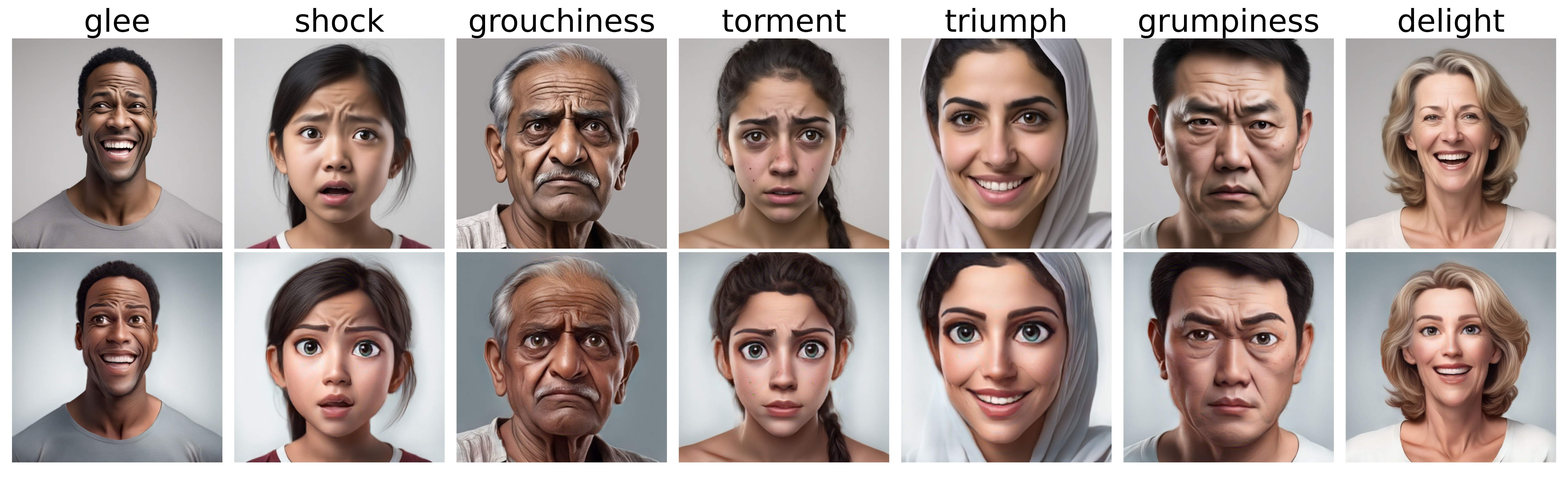}\\
    \rotatebox{90}{\hspace{0.7cm}Cartoon\hspace{1cm} Realistic}
    \includegraphics[width=0.95\textwidth]{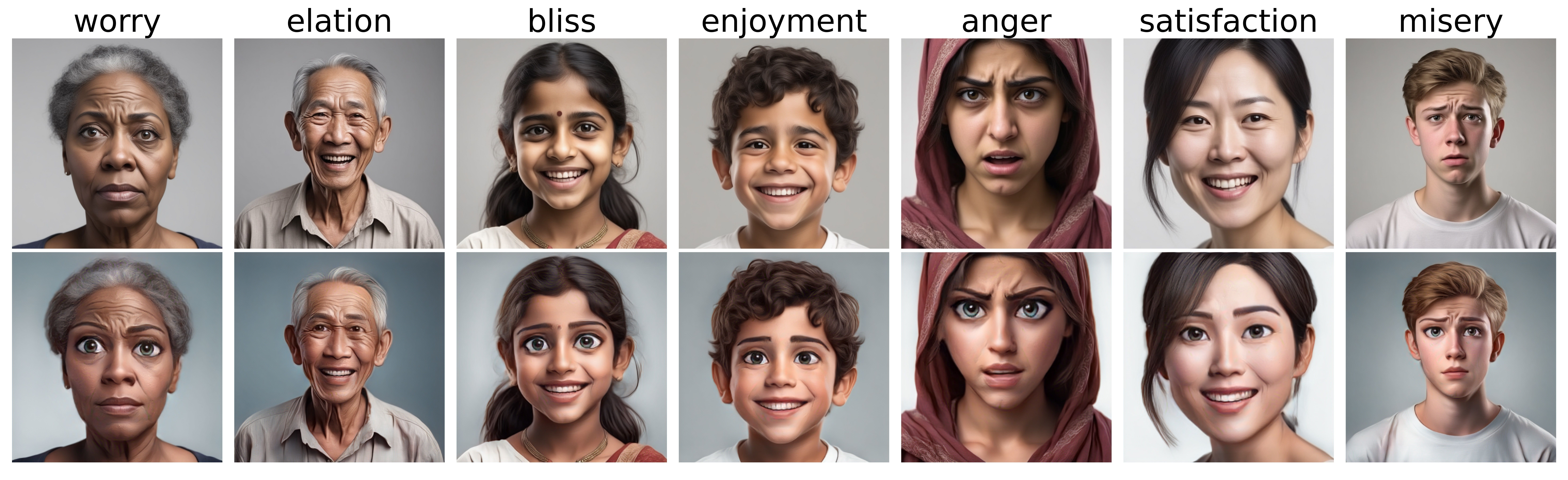}\\
    \rotatebox{90}{\hspace{0.7cm}Cartoon\hspace{1cm} Realistic}
    \includegraphics[width=0.95\textwidth]{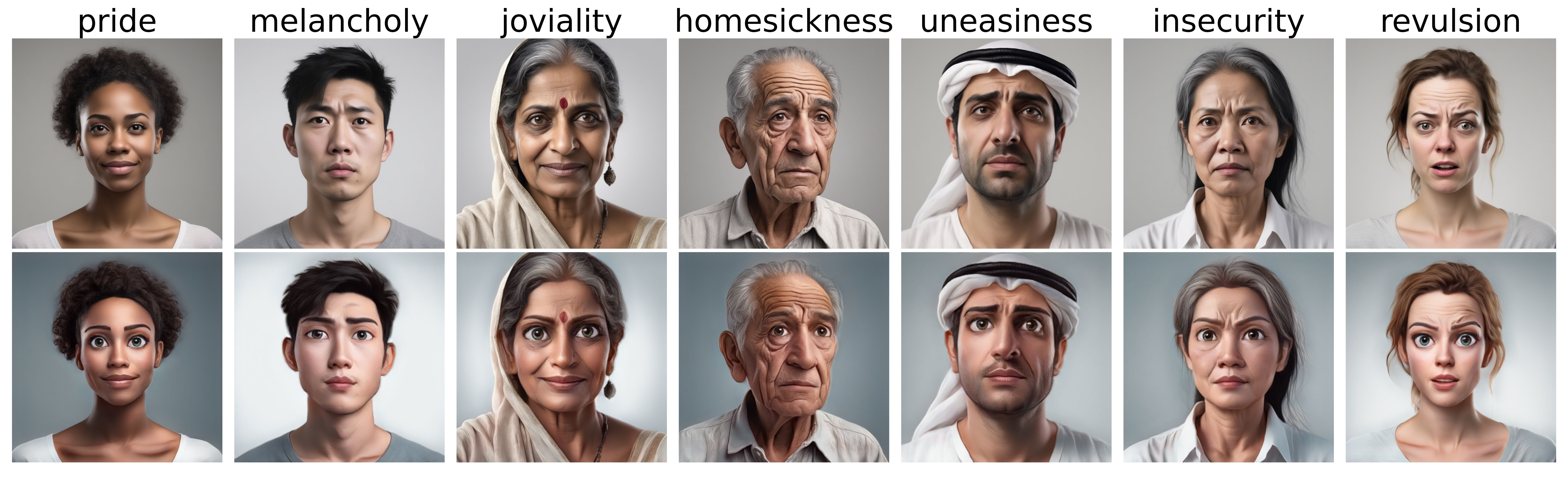}
    \caption{Examples of realistic and stylized images across a variety of facial expressions in GenEAva 1.0. The images illustrate diverse age groups and a balanced representation of race and gender. The stylization effectively preserves the identity and expressions of the realistic images.}
    \label{fig:sample_cartoons}
\end{figure*}

\section{Dataset: GenEAva 1.0}

Building on the pipeline described above, we present GenEAva 1.0, a novel dataset of \textbf{Gen}erative \textbf{E}xpressive \textbf{Ava}tars. GenEAva 1.0 comprises 13,230 cartoon avatars of 135 facial expressions. Each facial expression is represented by 98 images, ensuring a balanced distribution across genders (male and female), seven racial groups, and various age ranges. 

Examples of generated photo-realistic faces and their corresponding cartoon avatars are shown in Figure~\ref{fig:sample_cartoons}. As illustrated in the figure with a sample of 21 different expressions, the dataset includes cartoon avatars with fine-grained facial expressions (135 classes) and represents diverse races, ages, and genders. The images also feature clean backgrounds, further enhancing their utility.

\section{Experiments}\label{sec:experiments}

\subsection{Experimental Setup} We conducted experiments for both facial expression generation and stylization to evaluate different aspects of the proposed framework. First, we assessed the performance of the facial expression generation model in producing images with accurate and fine-grained facial expressions. Second, we investigated potential memorization issues within the model to ensure it generates images with unseen identities and avoids replicating those present in the fine-tuning data. Lastly, we evaluated the stylization model to determine its ability to preserve the original content of the images, focusing specifically on maintaining identity and facial expressions during the stylization process.

\subsection{Facial Expression Generation}

\begin{table}
    \caption{Facial expression generation results. Our model outperforms the SDXL model across all metrics, including CLIP, DINO, LPIPS, and Expression Error (Exp.)\ scores.}
    \label{table:exp_generation}
    \centering
    \normalsize
    \begin{tabular}{l|cccc}
        \hline
         Model &  CLIP$\uparrow$ & DINO$\uparrow$ & LPIPS$\downarrow$ & Exp.$\downarrow$ \\
        \hline
         SDXL~\cite{podell2023sdxl} & 0.780 & 0.738 & 0.658 & 13.1 \\
         Ours & \textbf{0.799} & \textbf{0.742} & \textbf{0.648} & \textbf{12.6} \\
        \hline
    \end{tabular}
\end{table}

We fine-tuned the SDXL model on the Emo135 dataset using LoRA~\cite{hu2021lora} with a rank of 4. We used the official SDXL checkpoint from Hugging Face.\footnote{https://huggingface.co/stabilityai/stable-diffusion-xl-base-1.0} The learning rate is set to 1e-6. We trained the model for eight epochs with a batch size of 1. More epochs lead to overfitting and worse image quality. The Adam optimizer~\cite{diederik2014adam} was used with $\beta_1=0.9, \beta_2=0.999$, and a weight decay of 1e-2. The expression loss weight $\alpha$ is set to 1.0. All experiments were conducted on four NVIDIA RTX A6000 GPUs.

To evaluate the model's ability to generate fine-grained facial expressions, we conducted experiments comparing our model to SDXL, the state-of-the-art T2I diffusion model. A total of 13,230 images were generated using SDXL with the same prompts used to create our proposed dataset GenEAva. We then computed multiple metrics to assess the fidelity and representation of facial expressions in both the SDXL-generated images and our dataset. For evaluation, we randomly sampled an evaluation subset from Emo135, consisting of 50 images for each facial expression. 

The following are the four metrics used for the discussed assessment: CLIP~\cite{radford2021learning}, DINO~\cite{caron2021emerging}, LPIPS~\cite{zhang2018perceptual}, and expression error (Exp.). The CLIP metric measures the average pairwise cosine similarity between the CLIP~\cite{radford2021learning} embeddings of the generated images and the Emo135 evaluation images, capturing semantic consistency between the generated and real facial expressions. The DINO metric is the average pairwise cosine similarity between the DINOv2~\cite{oquab2023dinov2} embeddings of the generated and evaluated images. LPIPS quantifies low-level perceptual differences, focusing on fine-grained texture and feature fidelity in images. It is calculated as the average pairwise similarity between the AlexNet~\cite{krizhevsky2012imagenet} activations of the generated and evaluation images. Finally, the expression error is the average pairwise Euclidean distance between the expression embeddings~\cite{zheng2023poster} of the generated and the evaluation images. We cropped the face in the image using RetinaFace~\cite{deng2019retinaface} and extracted the expression embedding using the facial expression model POSTER~\cite{zheng2023poster}.

The results of GenEAva and the baseline SDXL, measured in terms of the four metrics, are presented in Table~\ref{table:exp_generation}. Our model outperforms SDXL across all the evaluation metrics, indicating the GenEAva's superior ability to generate fine-grained facial expressions.  

\begin{figure*}[t]
    \centering
    \includegraphics[width=0.8\textwidth]{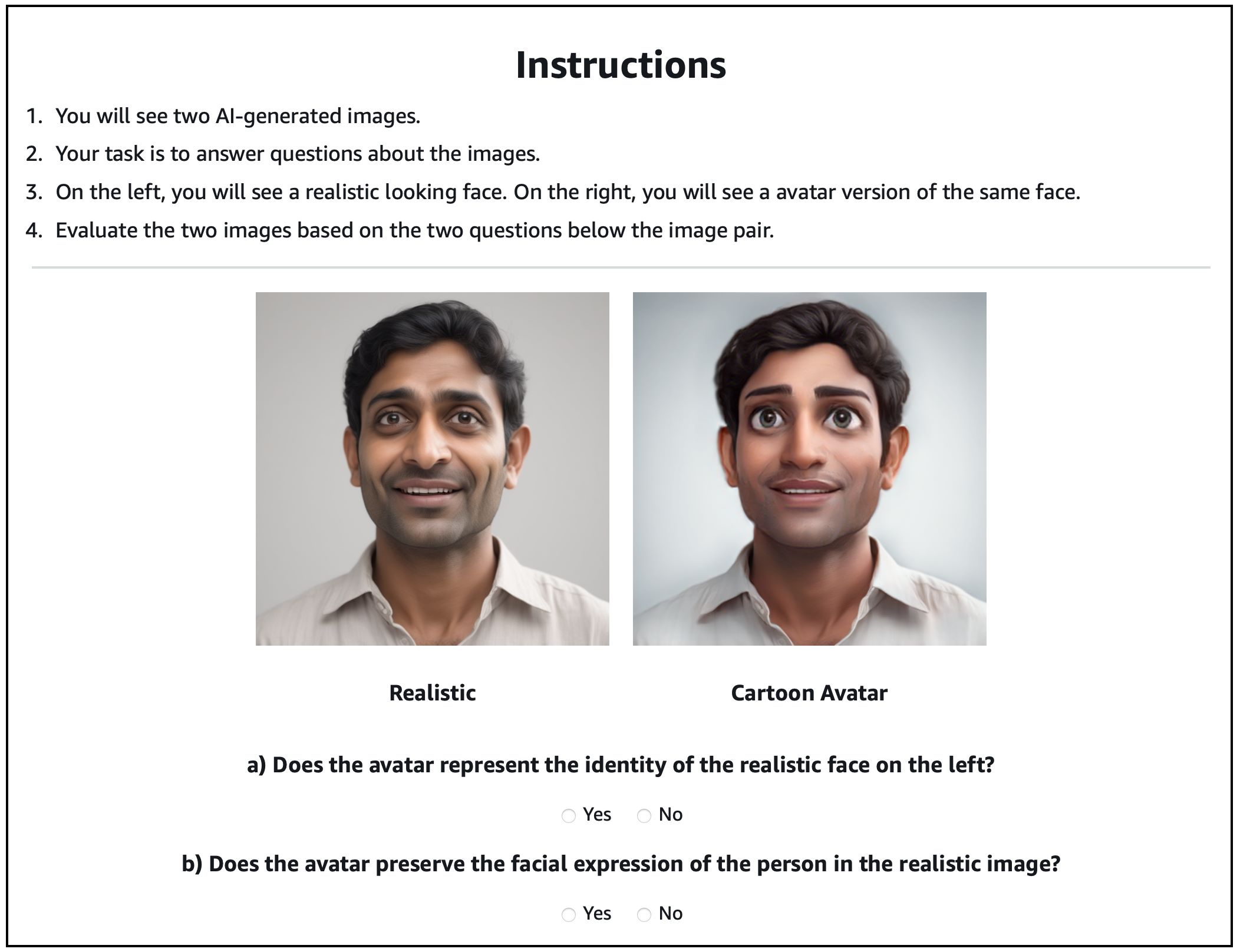}
    \caption{Interface for the Amazon Mechanical Turk (AMT) user study. The first question addresses the preservation of identity through the stylization module, and the second question addresses the preservation of the facial expression through the stylization module. Each evaluator was presented with 15 such examples, one of which is a test question presenting two images with obviously different identities. This is used to evaluate the validity of the HIT. Nine Turkers were recruited to complete each HIT. Invalid HITs were discarded.}
    \label{fig:user_study_interface}
\end{figure*}

\subsection{Identity Memorization in Facial Expression Generation}


To evaluate whether the trained facial expression generation model memorizes the identities in the fine-tuning data, we compared the identity embeddings of generated images in GenEAva 1.0 with those of the images in the fine-tuning dataset. First, we used RetinaFace~\cite{deng2019retinaface} to extract faces from images and a face recognition model, ArcFace~\cite{deng2019arcface}, to compute identity embeddings.
The cosine similarity between the identity embeddings of the generated images and the fine-tuning images is then calculated. For each generated image, we identified the training image with the highest cosine similarity (the most similar face) and determined whether the two images belong to the same identity on the basis of (1) an empirical verification threshold and (2) a user study. Using an algorithm that maximizes the information gain~\cite{serengil2020lightface}, the threshold is set to 0.68. If the cosine similarity was to exceed this threshold, the two faces would be considered to have the same identity. We also conducted a user study to check whether the generated images replicate identities from previously seen data. 

\subsubsection{Quantitative Analysis}
We first compared the generated images with those in our fine-tuning dataset Emo135~\cite{chen2022semantic}. None of the generated faces exceeded the verification similarity threshold~\cite{serengil2020lightface} for any of the faces in the Emo135 dataset. Additionally, the average pairwise cosine similarity between the generated faces and their most similar counterparts in the fine-tuning set is 0.39, which is significantly below the verification threshold. This suggests that the model did not memorize the identities from the fine-tuning dataset.

Ideally, we would also evaluate this metric on the entire SDXL training dataset. However, since the SDXL training data is not fully publicly accessible, we used CelebA~\cite{liu2018large}, a commonly used face dataset known to be part of SDXL's training data. There are 10,177 distinct identities in the dataset, and we randomly selected one image for each identity for comparison. None of the generated faces exceeds the verification similarity threshold when matched against any of the CelebA faces. Additionally, the average cosine similarity between the embeddings of the generated faces and their closest counterparts in CelebA is 0.47. 

\subsubsection{User Study}
To further validate that identities have not been memorized in the generated images, we conducted a user study.
We randomly sampled 50 images from GenEAva 1.0 and found their most similar faces in the fine-tuning dataset Emo135 based on identity similarity. Similarly, we also sampled 50 images for comparison with the CelebA dataset. 
%
%
Then, five participants were asked to determine whether the paired faces belong to the same person.
Out of 100 pairs, two participants identified one pair of faces as the same identity, two participants identified two pairs, and one participant identified five pairs.
We note that all these faces are from the CelebA dataset, and the pairs share similarities, but are not exact replications.
This suggests that our model did not memorize the identities in the fine-tuning dataset, Emo135.
Also, a few similar faces to CelebA could be attributed to SDXL's pretraining on large-scale public datasets, where celebrities are overrepresented.
We will remove these faces that are identified as similar to CelebA in the final version of our dataset.
We will also apply the latest unlearning algorithm (\textit{e.g.,}~\cite{li2024machine}) to make the model forget such content. 

Overall, these results demonstrate that our model does not memorize identities from the fine-tuning dataset, and that the generated dataset does not include any identities present in these training data. The proposed GenEAva 1.0 establishes a benchmark with new identities that feature a balanced representation of race, gender, and age, promoting both privacy and fairness.

\subsection{Identity and Facial Expression Preservation in Face Stylization}


Finally, we evaluate the performance of face stylization. An ideal stylization method should transform the image into the desired style while preserving the original content that is not related to style. For avatar generation, we specifically focus on maintaining the identity and facial expression of the face. This evaluation was carried out with a user study.

\subsubsection{User Study}\label{sec:style_userstudy}

\begin{figure}[t!]
    \centering
    \includegraphics[width=\columnwidth]{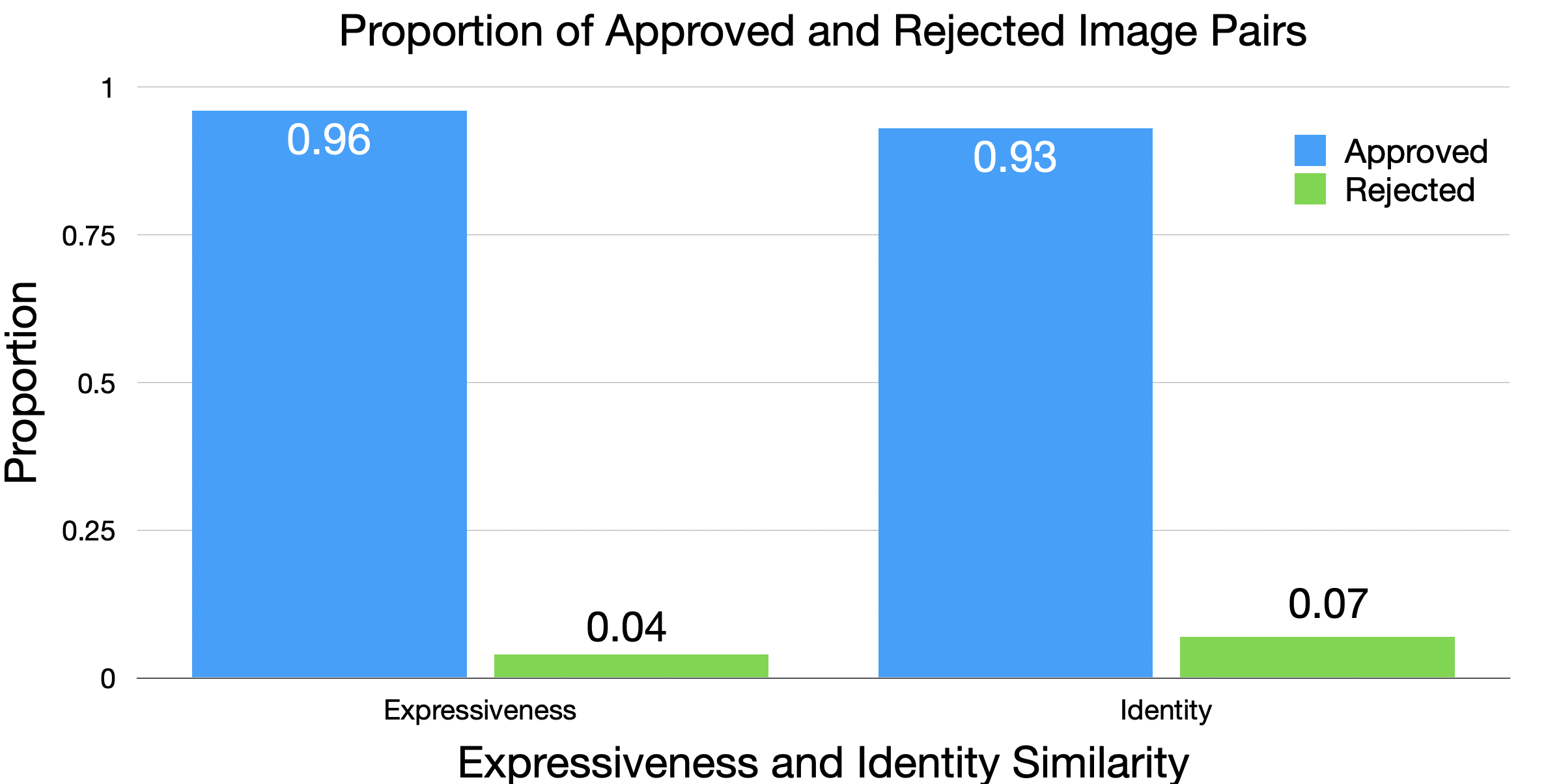}
    \caption{User study results evaluating the stylization based on identity preservation and expression preservation. We achieved 96\% approval rating in preserving facial expression and 93\% approval rating in preserving identity, indicating the effectiveness of the stylization method. The approval rating indicates the percentage  (\%) of pairs that preserve the facial expression and identity among them.  }
    \label{fig:user_study_curve_2}
\end{figure}

We conducted a user study comparing the realistic fine-grained facial expression image to its corresponding avatar generation in terms of (i) identity preservation, and (ii) facial expression preservation. We used the Amazon Mechanical Turk (AMT) crowdsourcing marketplace\footnote{https://www.mturk.com/} to recruit seven crowd workers for each task. For the user study, we randomly sampled 945 generated image-avatar pairs (seven from each of 135 emotion classes).
We recruited AMT workers who had previously completed at least 500 tasks (`HITs'), and maintained an approval rating of at least 95\%. We compensated for the work of all crowd workers who participated in our tasks. For each HIT, a total of 9 evaluators were assigned. 
Each subtask presents the worker with one realistic fine-grained expression image and its corresponding avatar image (Figure~\ref{fig:user_study_interface}).  The worker is asked to determine whether the identity and fine-grained facial expression are maintained from the realistic generation to the resulting avatar. We posted all HITs simultaneously and allocated a maximum of 9 minutes to complete each HIT. Each evaluator was presented with 15 image pairs, one of which is a test question presenting two images with obviously different identities. This is used to evaluate the validity of the HIT response. Invalid HITs, i.e., those with an evaluator incorrectly answering the test questions, were discarded. 

The results show that our dataset achieved 96\% approval rating in facial expression preservation and 93\% approval rating in identity preservation (Figure~\ref{fig:user_study_curve_2}). 
This suggests that the stylization method effectively maintains the facial expression and identity of the images.
We will remove images that fail to preserve the identity and facial expression from the final version of the dataset. 

\begin{figure*}[t]
    \centering
    Compassion \hspace{1.1cm} Desire    \hspace{1.5cm}  Jealousy \hspace{1.2cm} Sympathy \\
    ~ \\
    \rotatebox{90}{\hspace{0.6cm}ChatGPT}
    \includegraphics[width=0.15\textwidth]{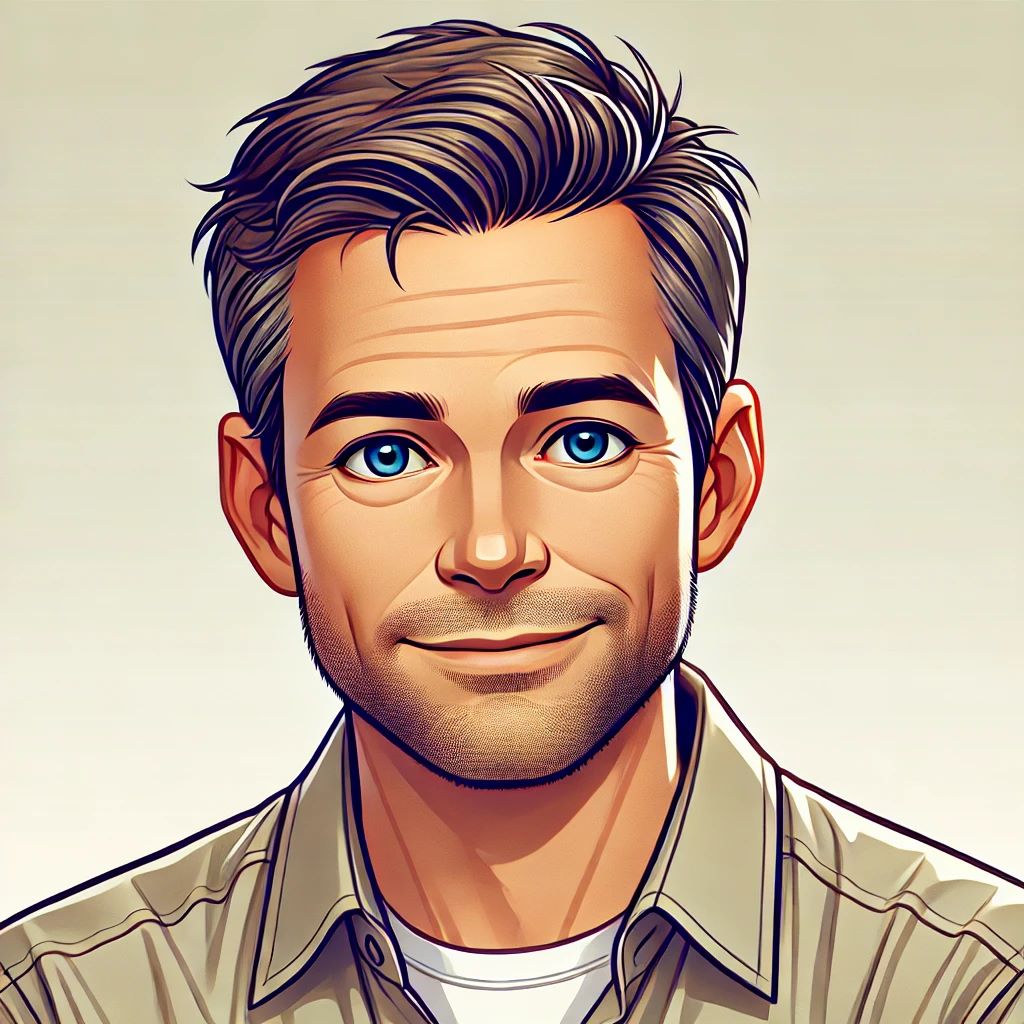}
    \includegraphics[width=0.15\textwidth]{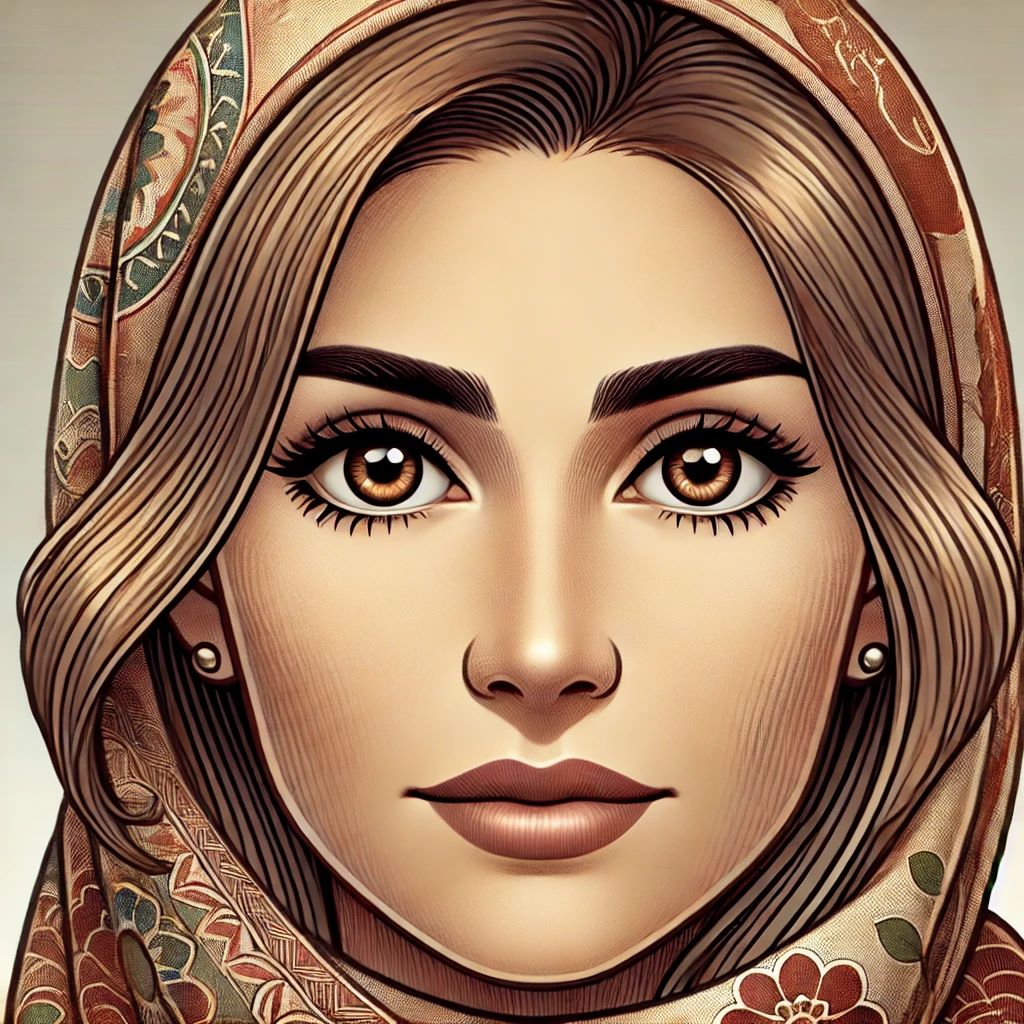}
    \includegraphics[width=0.15\textwidth]{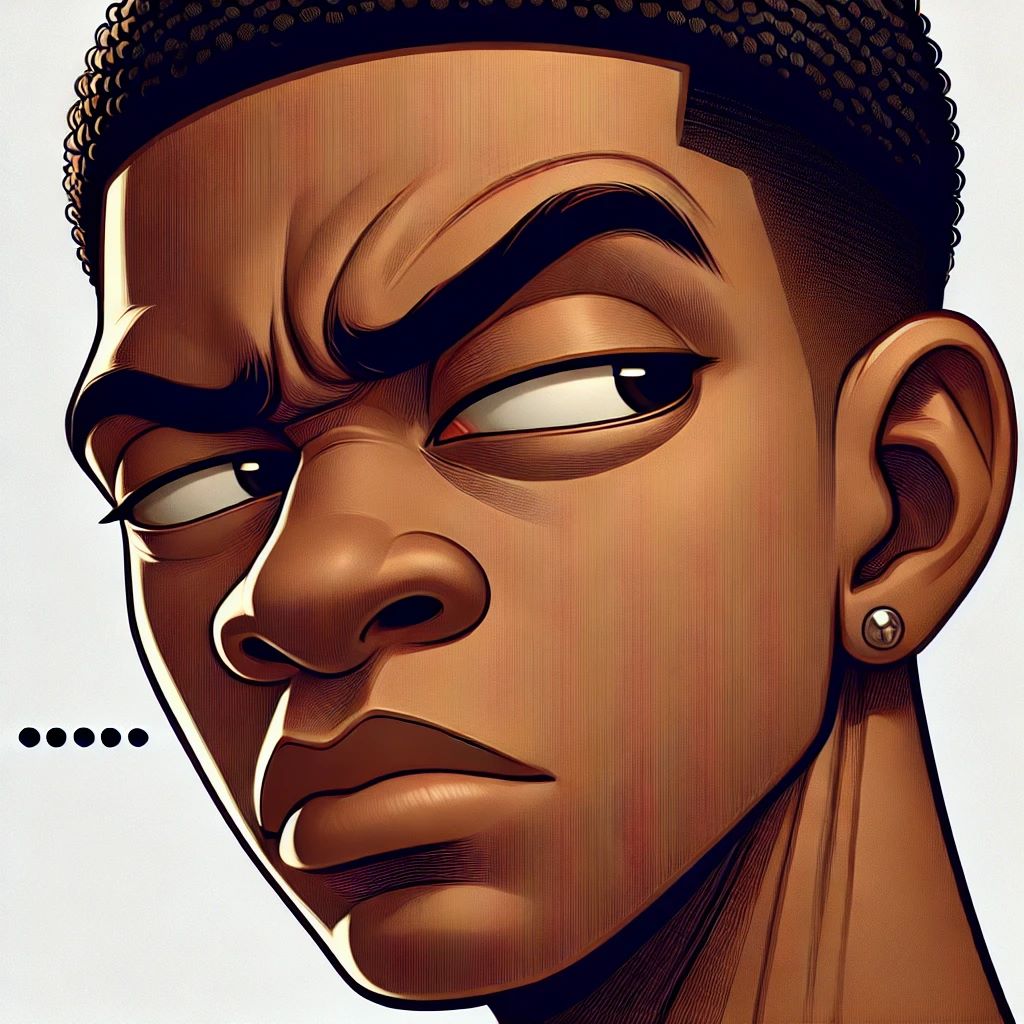}
    \includegraphics[width=0.15\textwidth]{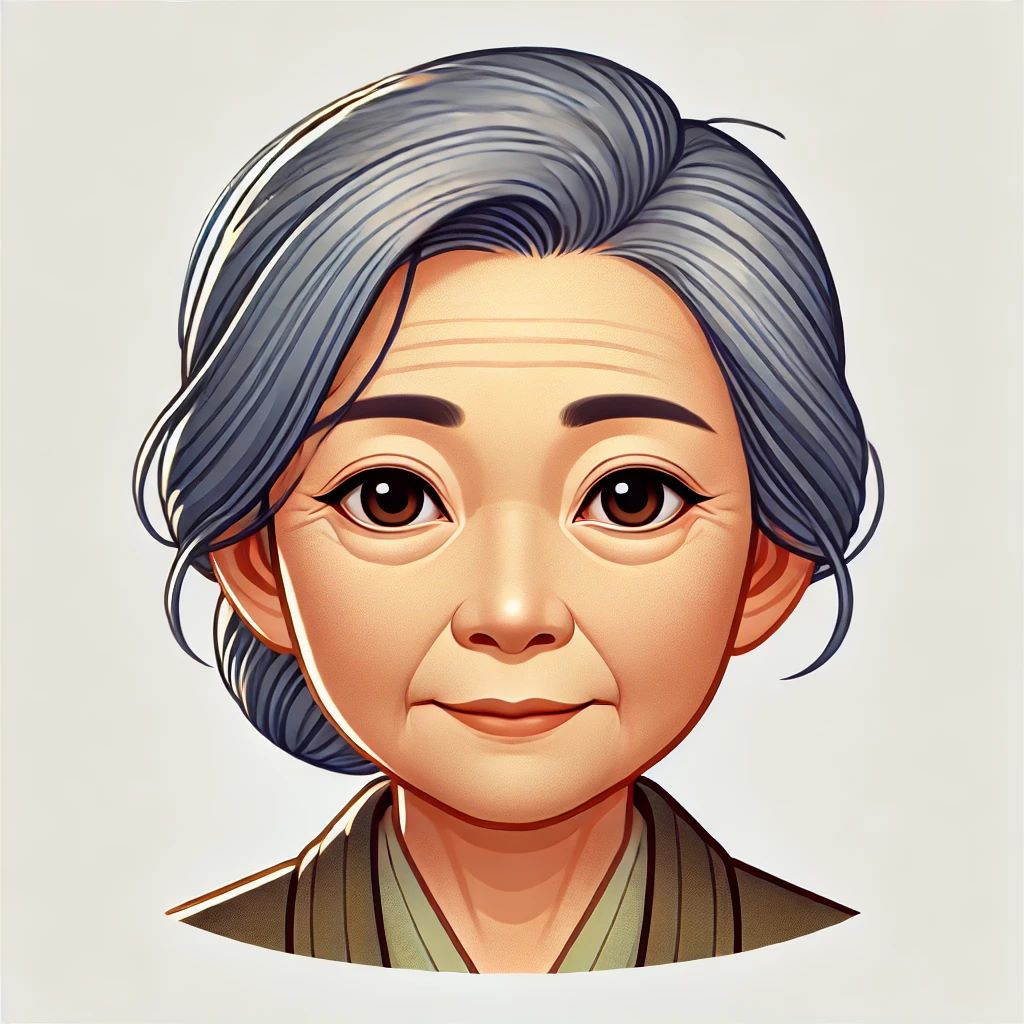}\\
    \vspace{0.15cm}
    \rotatebox{90}{\hspace{0.6cm}GenEAva}
    \includegraphics[width=0.15\textwidth]{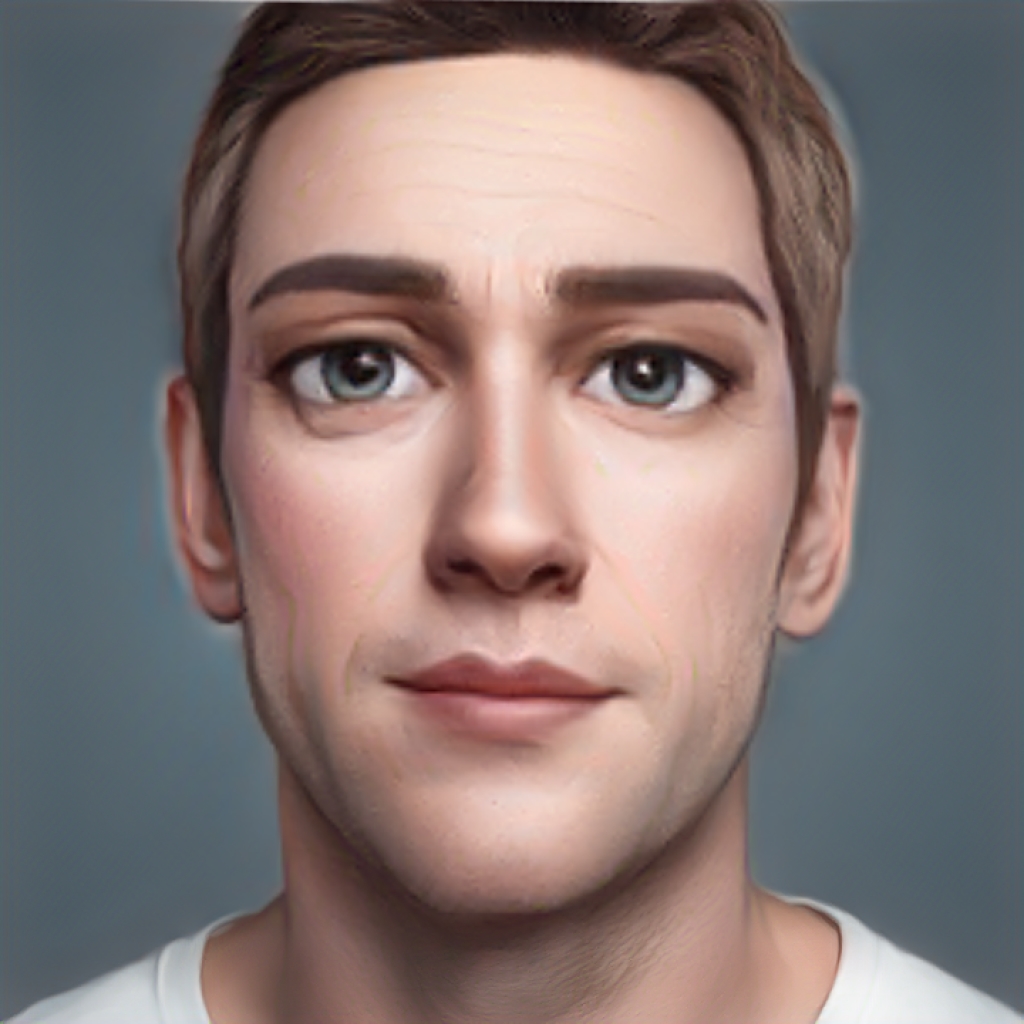}
    \includegraphics[width=0.15\textwidth]{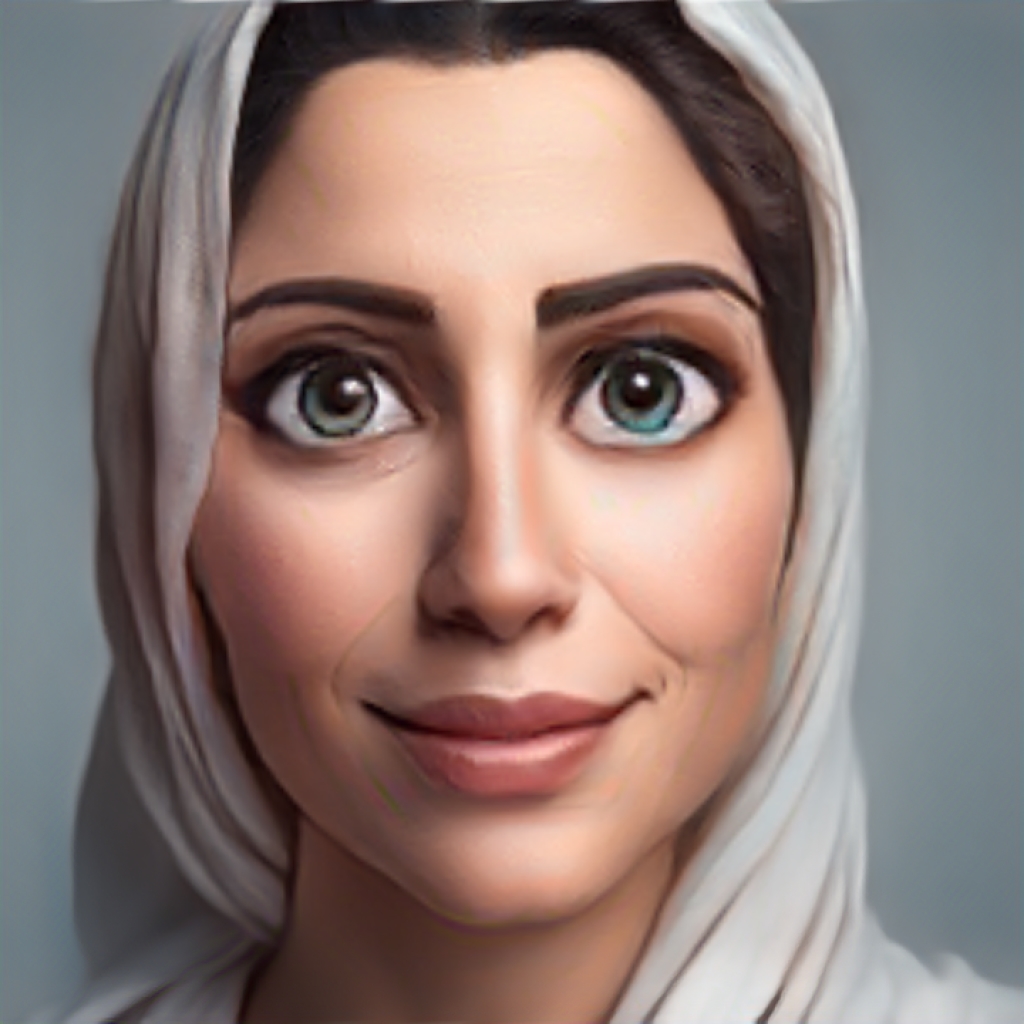}
    \includegraphics[width=0.15\textwidth]{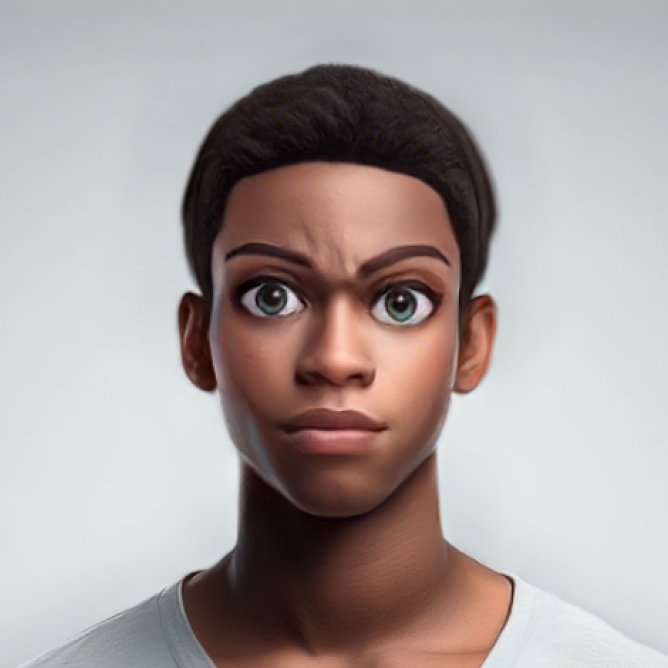}
    \includegraphics[width=0.15\textwidth]{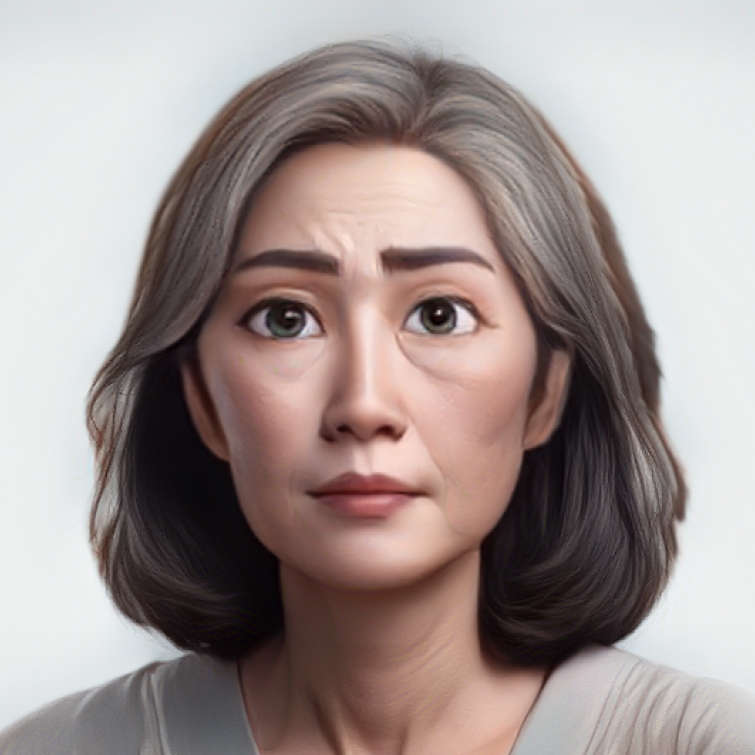}
    \caption{Qualitative examples of images generated by ChatGPT~\cite{openai2024chatgpt} and our proposed GenEAva. GenEAva shows a superior ability to capture subtle expressions compared to ChatGPT, which either produces generic neutral or exaggerated expressions. }
    \label{fig:qualitative}
\end{figure*}

\subsection{Qualitative Results}
We present qualitative examples that compare our cartoon avatars with those generated by ChatGPT~\cite{openai2024chatgpt} using GPT-4o~\cite{hurst2024gpt} and DALLE-3~\cite{openai2024dalle3} in Figure~\ref{fig:qualitative}. 
For ChatGPT-generated images, the sample prompt we used is: \textit{``Generate a detailed cartoon avatar of a middle-aged White male showing a facial expression of compassion.''}

As shown in Figure~\ref{fig:qualitative}, our proposed method GenEAva captures subtle facial expressions more accurately than ChatGPT.
For example, `compassion' is defined as sympathetic pity and concern for the sufferings or misfortunes of others. However, the face generated by ChatGPT depicts a smile while GenEAva produces a slight frown, which better aligns with the intended expression. 
For some classes like `desire' and `sympathy,' ChatGPT generates neutral faces, while our proposed method correctly captures the subtle expressions.
For `jealousy,' the ChatGPT-generated face is overly exaggerated compared to the proposed GenEAva.
Overall, we show that it is challenging even for commercial multimodel LLMs like ChatGPT to accurately generate certain subtle facial expressions. Our proposed method is able to successfully create faces with fine-grained facial expressions.

\section{CONCLUSIONS}

In this work, we presented a novel framework, GenEAva, which can generate and validate high-quality cartoon avatars with fine-grained facial expressions while ensuring unique identities and diversity in age, gender, and racial representation.
We fine-tuned a text-to-image diffusion model with expression-guided objectives to generate highly detailed and expressive facial expressions. 
We designed comprehensive prompts for the T2I diffusion model to produce images with equal representation of males and females, a wide range of ages from teenagers to elderly individuals, and a balanced representation across seven racial groups (White, Black, Indian, East Asian, Southeast Asian, Middle
East, and Latino).
Finally, we applied a stylization model that converts these realistic faces into cartoon avatars while preserving both identity and expression.
In addition, we presented an evaluation pipeline for cartoon avatar generation, assessing the fidelity and representation of facial expression, identity memorization, and the preservation of identity and expression in stylization.

Leveraging this framework, we introduced GenEAva 1.0, the first expressive cartoon avatar dataset specifically designed to capture 135 fine-grained facial expressions, featuring 13,230 cartoon avatars with a balanced distribution across demographic attributes.
Extensive experiments demonstrate that our model outperforms state-of-the-art diffusion-based text-to-image models, such as SDXL, in generating subtle and accurate facial expressions. 
We verified that the generated identities are novel and not memorized from the training set through both quantitative analysis and a user study.
We also showed that identity and fine-grained facial expressions are preserved by the stylization method. 

Our findings highlight the potential of diffusion-based models in advancing expressive cartoon avatar generation. 
Future work could explore further improving expression control, improving identity consistency across different expressions, and adapting the framework for real-time applications.





\section*{ETHICAL IMPACT STATEMENT}

Our work aims at generating cartoon avatars with fine-grained facial expressions. 
As with all generative models, our model has inherent risks and potential negative impacts, including potential misuse for generating misleading or harmful content, privacy concerns regarding the unauthorized use of individuals’ likenesses, and the risk of reinforcing or introducing biases through the generated images.
Furthermore, while facial expression generation and manipulation could enhance digital interactions, gaming, and virtual communication, it also poses ethical concerns, such as impersonation and deceptive emotional representations.
We acknowledge these risks and emphasize the need for responsible use, transparency, and ethical deployment of the proposed model and AI-generated content.
To mitigate these risks, we ensure that there is no identity memorization in our model and dataset through extensive experiments.
We also firmly oppose the use of our model for generating fake or misleading content.
Future users of our dataset will be required to agree to our terms of use and ethical guidelines.



\bibliographystyle{ieee}
\bibliography{egbib}

\begin{thebibliography}{10}\itemsep=-1pt

\bibitem{abbasnejad2017using}
I.~Abbasnejad, S.~Sridharan, D.~Nguyen, S.~Denman, C.~Fookes, and S.~Lucey.
\newblock Using synthetic data to improve facial expression analysis with 3d convolutional networks.
\newblock In {\em Proceedings of the IEEE International Conference on Computer Vision Workshops}, pages 1609--1618, 2017.

\bibitem{azari2024emostyle}
B.~Azari and A.~Lim.
\newblock Emo{S}tyle: {}ne-shot facial expression editing using continuous emotion parameters.
\newblock In {\em Proceedings of the IEEE/CVF Winter Conference on Applications of Computer Vision}, pages 6385--6394, 2024.

\bibitem{bae2023digiface1m}
G.~Bae, M.~de~La~Gorce, T.~Baltru{\v{s}}aitis, C.~Hewitt, D.~Chen, J.~Valentin, R.~Cipolla, and J.~Shen.
\newblock Digi{F}ace-1{M}: 1 million digital face images for face recognition.
\newblock In {\em 2023 IEEE Winter Conference on Applications of Computer Vision (WACV)}. IEEE, 2023.

\bibitem{bie2023private}
A.~Bie, G.~Kamath, and G.~Zhang.
\newblock Private {GAN}s, revisited.
\newblock {\em Transactions on Machine Learning Research}, 2023.
\newblock Survey Certification.

\bibitem{bimberg2024influence}
P.~Bimberg, M.~Feldmann, B.~Weyers, and D.~Zielasko.
\newblock The influence of environmental context on the creation of cartoon-like avatars in virtual reality.
\newblock In {\em 2024 IEEE Conference Virtual Reality and 3D User Interfaces (VR)}, pages 298--308. IEEE, 2024.

\bibitem{Sface_Boutros}
F.~Boutros, M.~Huber, P.~Siebke, T.~Rieber, and N.~Damer.
\newblock {SFace: P}rivacy-friendly and accurate face recognition using synthetic data.
\newblock In {\em {IEEE} International Joint Conference on Biometrics, {IJCB} 2022, Abu Dhabi, United Arab Emirates, October 10-13, 2022}, pages 1--11. {IEEE}, 2022.

\bibitem{branwen2019danbooru2019}
G.~Branwen and A.~Gokaslan.
\newblock Danbooru2019: A large-scale crowdsourced and tagged anime illustration dataset.
\newblock https://gwern.net/danbooru2021\#danbooru2019, 2019.

\bibitem{carlini_usenix}
N.~Carlini, J.~Hayes, M.~Nasr, M.~Jagielski, V.~Sehwag, F.~Tram\`{e}r, B.~Balle, D.~Ippolito, and E.~Wallace.
\newblock Extracting training data from diffusion models.
\newblock In {\em Proceedings of the 32nd USENIX Conference on Security Symposium}, SEC '23, USA, 2023. USENIX Association.

\bibitem{caron2021emerging}
M.~Caron, H.~Touvron, I.~Misra, H.~J\'egou, J.~Mairal, P.~Bojanowski, and A.~Joulin.
\newblock Emerging properties in self-supervised vision transformers.
\newblock In {\em Proceedings of the International Conference on Computer Vision (ICCV)}, 2021.

\bibitem{chen_privacy}
C.~Chen, D.~Liu, S.~Ma, S.~Nepal, and C.~Xu.
\newblock { Private Image Generation with Dual-Purpose Auxiliary Classifier }.
\newblock In {\em 2023 IEEE/CVF Conference on Computer Vision and Pattern Recognition (CVPR)}, pages 20361--20370, Los Alamitos, CA, USA, June 2023. IEEE Computer Society.

\bibitem{chen_memorisation}
C.~Chen, D.~Liu, and C.~Xu.
\newblock Towards memorization-free diffusion models.
\newblock In {\em 2024 IEEE/CVF Conference on Computer Vision and Pattern Recognition (CVPR)}, pages 8425--8434, 2024.

\bibitem{chen2022semantic}
K.~Chen, X.~Yang, C.~Fan, W.~Zhang, and Y.~Ding.
\newblock Semantic-rich facial emotional expression recognition.
\newblock {\em IEEE Transactions on Affective Computing}, 13(4):1906--1916, 2022.

\bibitem{choi2018stargan}
Y.~Choi, M.~Choi, M.~Kim, J.-W. Ha, S.~Kim, and J.~Choo.
\newblock Star{GAN: U}nified generative adversarial networks for multi-domain image-to-image translation.
\newblock In {\em Proceedings of the IEEE Conference on Computer Vision and Pattern Recognition}, pages 8789--8797, 2018.

\bibitem{d2021ganmut}
S.~d'Apolito, D.~P. Paudel, Z.~Huang, A.~Romero, and L.~Van~Gool.
\newblock {GAN}mut: {L}earning interpretable conditional space for gamut of emotions.
\newblock In {\em Proceedings of the IEEE/CVF Conference on Computer Vision and Pattern Recognition}, pages 568--577, 2021.

\bibitem{deng2019arcface}
J.~Deng, J.~Guo, N.~Xue, and S.~Zafeiriou.
\newblock Arc{F}ace: {A}dditive angular margin loss for deep face recognition.
\newblock In {\em Proceedings of the IEEE/CVF Conference on Computer Vision and Pattern Recognition}, pages 4690--4699, 2019.

\bibitem{deng2019retinaface}
J.~Deng, J.~Guo, Y.~Zhou, J.~Yu, I.~Kotsia, and S.~Zafeiriou.
\newblock Retina{F}ace: {S}ingle-stage dense face localisation in the wild.
\newblock {\em arXiv preprint arXiv:1905.00641}, May 2019.

\bibitem{dockhorn2023differentially}
T.~Dockhorn, T.~Cao, A.~Vahdat, and K.~Kreis.
\newblock Differentially private diffusion models.
\newblock {\em Transactions on Machine Learning Research}, 2023.

\bibitem{fink2024ai}
M.~C. Fink, S.~A. Robinson, and B.~Ertl.
\newblock {AI}-based avatars are changing the way we learn and teach: benefits and challenges.
\newblock In {\em Frontiers in Education}, volume~9, page 1416307. Frontiers Media SA, 2024.

\bibitem{fujimoto2016manga109}
A.~Fujimoto, T.~Ogawa, K.~Yamamoto, Y.~Matsui, T.~Yamasaki, and K.~Aizawa.
\newblock Manga109 dataset and creation of metadata.
\newblock In {\em MANPU '16: Proceedings of the 1st International Workshop on coMics Analysis, Processing and Understanding}, pages 1--5, 2016.

\bibitem{goodfellow2020generative}
I.~Goodfellow, J.~Pouget-Abadie, M.~Mirza, B.~Xu, D.~Warde-Farley, S.~Ozair, A.~Courville, and Y.~Bengio.
\newblock Generative adversarial networks.
\newblock {\em Communications of the ACM}, 63(11):139--144, 2020.

\bibitem{cartoonset}
Google.
\newblock Cartoon{S}et: {A} dataset of cartoon faces.
\newblock \url{https://google.github.io/cartoonset/}, 2021.
\newblock Accessed: 2025-01-18.

\bibitem{hadjiev2021evaluation}
A.~M. Hadjiev and K.~Araki.
\newblock Evaluation of various avatar designs for conversational chatbot systems.
\newblock {\em SIG-LSE JSAI}, 2021.

\bibitem{he2024synfer}
X.~He, C.~Luo, X.~Xian, B.~Li, S.~Song, M.~H. Khan, W.~Xie, L.~Shen, and Z.~Ge.
\newblock Syn{FER: T}owards boosting facial expression recognition with synthetic data.
\newblock {\em arXiv preprint arXiv:2410.09865}, 2024.

\bibitem{ho2020denoising}
J.~Ho, A.~Jain, and P.~Abbeel.
\newblock Denoising diffusion probabilistic models.
\newblock {\em Advances in Neural Information Processing Systems}, 33:6840--6851, 2020.

\bibitem{hu2021lora}
E.~J. Hu, Y.~Shen, P.~Wallis, Z.~Allen-Zhu, Y.~Li, S.~Wang, L.~Wang, and W.~Chen.
\newblock Lo{RA: L}ow-rank adaptation of large language models.
\newblock {\em arXiv preprint arXiv:2106.09685}, 2021.

\bibitem{huoa2017webcaricature}
J.~Huoa, W.~Lia, Y.~Shia, Y.~Gaoa, and H.~Yinb.
\newblock Webcaricature: a benchmark for caricature face recognition.
\newblock {\em arXiv preprint arXiv:1703.03230}, 2017.

\bibitem{hurst2024gpt}
A.~Hurst, A.~Lerer, A.~P. Goucher, A.~Perelman, A.~Ramesh, A.~Clark, A.~Ostrow, A.~Welihinda, A.~Hayes, A.~Radford, et~al.
\newblock G{PT}-4o {S}ystem {C}ard.
\newblock {\em arXiv preprint arXiv:2410.21276}, Oct. 2024.
\newblock 33 pages.

\bibitem{karkkainen2021fairface}
K.~Karkkainen and J.~Joo.
\newblock Fair{F}ace: {F}ace attribute dataset for balanced race, gender, and age for bias measurement and mitigation.
\newblock In {\em Proceedings of the IEEE/CVF Winter Conference on Applications of Computer Vision (WACV)}, pages 1548--1558, 2021.

\bibitem{karras2020analyzing}
T.~Karras, S.~Laine, M.~Aittala, J.~Hellsten, J.~Lehtinen, and T.~Aila.
\newblock Analyzing and improving the image quality of {S}tyle{GAN}.
\newblock In {\em Proceedings of the IEEE/CVF Conference on Computer Vision and Pattern Recognition}, pages 8110--8119, 2020.

\bibitem{kim2023dcface}
M.~Kim, F.~Liu, A.~Jain, and X.~Liu.
\newblock {DCFace: S}ynthetic face generation with dual condition diffusion model.
\newblock In {\em Proceedings of the IEEE/CVF Conference on Computer Vision and Pattern Recognition}, pages 12715--12725, 2023.

\bibitem{diederik2014adam}
D.~P. Kingma and J.~Ba.
\newblock Adam: {A} method for stochastic optimization.
\newblock {\em arXiv preprint arXiv:1412.6980}, 2014.

\bibitem{krizhevsky2012imagenet}
A.~Krizhevsky, I.~Sutskever, and G.~E. Hinton.
\newblock Image{N}et classification with deep convolutional neural networks.
\newblock {\em Advances in neural information processing systems}, 25, 2012.

\bibitem{li2024machine}
G.~Li, H.~Hsu, C.-F. Chen, and R.~Marculescu.
\newblock Machine unlearning for image-to-image generative models.
\newblock In {\em The Twelfth International Conference on Learning Representations}, 2024.

\bibitem{liu2024towards}
R.~Liu, B.~Ma, W.~Zhang, Z.~Hu, C.~Fan, T.~Lv, Y.~Ding, and X.~Cheng.
\newblock Towards a simultaneous and granular identity-expression control in personalized face generation.
\newblock In {\em Proceedings of the IEEE/CVF Conference on Computer Vision and Pattern Recognition}, pages 2114--2123, 2024.

\bibitem{liu2018large}
Z.~Liu, P.~Luo, X.~Wang, and X.~Tang.
\newblock Large-scale {C}elebfaces attributes ({C}eleb{A}) dataset.
\newblock {\em Retrieved August}, 15(2018):11, 2018.

\bibitem{melzi2023gandiffface}
P.~Melzi, C.~Rathgeb, R.~Tolosana, R.~Vera-Rodriguez, D.~Lawatsch, F.~Domin, and M.~Schaubert.
\newblock {GANDiffFace: C}ontrollable generation of synthetic datasets for face recognition with realistic variations.
\newblock In {\em Proceedings of the IEEE/CVF International Conference on Computer Vision}, pages 3086--3095, 2023.

\bibitem{men2022dct}
Y.~Men, Y.~Yao, M.~Cui, Z.~Lian, and X.~Xie.
\newblock {DCT-Net: D}omain-calibrated translation for portrait stylization.
\newblock {\em ACM Transactions on Graphics (TOG)}, 41(4):1--9, 2022.

\bibitem{mishra2016iiit}
A.~Mishra, S.~N. Rai, A.~Mishra, and C.~Jawahar.
\newblock {IIIT-CFW: A} benchmark database of cartoon faces in the wild.
\newblock In {\em Computer Vision--ECCV 2016 Workshops: Amsterdam, The Netherlands, October 8-10 and 15-16, 2016, Proceedings, Part I 14}, pages 35--47. Springer, 2016.

\bibitem{mishra2023enhancing}
S.~Mishra, P.~Shalu, and R.~Singh.
\newblock Enhancing face emotion recognition with facs-based synthetic dataset using deep learning models.
\newblock In {\em International Conference on Computer Vision and Image Processing}, pages 523--531. Springer, 2023.

\bibitem{nizam2022avatar}
D.~N.~M. Nizam, D.~N. Rudiyansah, N.~M. Tuah, Z.~H.~A. Sani, and K.~Sungkaew.
\newblock Avatar design types and user engagement in digital educational games during evaluation phase.
\newblock {\em International Journal of Electrical and Computer Engineering}, 12(6):6449, 2022.

\bibitem{nowak2018avatars}
K.~L. Nowak and J.~Fox.
\newblock Avatars and computer-mediated communication: {A} review of the definitions, uses, and effects of digital representations.
\newblock {\em Review of Communication Research}, 6:30--53, 2018.

\bibitem{openai2024chatgpt}
OpenAI.
\newblock Chat{GPT}: {C}onversational {AI by OpenAI}.
\newblock OpenAI: https://openai.com/chatgpt, 2024.

\bibitem{openai2024dalle3}
OpenAI.
\newblock {DALL·E} 3: {O}pen{AI}'s text-to-image model.
\newblock OpenAI: https://openai.com/dall-e, 2024.

\bibitem{oquab2023dinov2}
M.~Oquab, T.~Darcet, T.~Moutakanni, H.~Vo, M.~Szafraniec, V.~Khalidov, P.~Fernandez, D.~Haziza, F.~Massa, A.~El-Nouby, et~al.
\newblock {DINOv2: L}earning robust visual features without supervision.
\newblock {\em arXiv preprint arXiv:2304.07193}, 2023.

\bibitem{panda2022alltogether}
P.~Panda, M.~J. Nicholas, M.~Gonzalez-Franco, K.~Inkpen, E.~Ofek, R.~Cutler, K.~Hinckley, and J.~Lanier.
\newblock All{T}ogether: {E}ffect of avatars in mixed-modality conferencing environments.
\newblock In {\em Proceedings of the 1st Annual Meeting of the Symposium on Human-Computer Interaction for Work}, pages 1--10, 2022.

\bibitem{watermark-detection}
I.~Pavlov.
\newblock Watermark detection, 2021.
\newblock GitHub repository: https://github.com/boomb0om/watermark-detection.

\bibitem{pikoulis2023photorealistic}
I.~Pikoulis, P.~P. Filntisis, and P.~Maragos.
\newblock Photorealistic and identity-preserving image-based emotion manipulation with latent diffusion models.
\newblock {\em arXiv preprint arXiv:2308.03183}, 2023.

\bibitem{podell2023sdxl}
D.~Podell, Z.~English, K.~Lacey, A.~Blattmann, T.~Dockhorn, J.~M{\"u}ller, J.~Penna, and R.~Rombach.
\newblock S{DXL: I}mproving latent diffusion models for high-resolution image synthesis.
\newblock {\em arXiv preprint arXiv:2307.01952}, 2023.

\bibitem{pumarola2018ganimation}
A.~Pumarola, A.~Agudo, A.~M. Martinez, A.~Sanfeliu, and F.~Moreno-Noguer.
\newblock {GAN}imation: {A}natomically-aware facial animation from a single image.
\newblock In {\em Proceedings of the European conference on Computer Vision (ECCV)}, pages 818--833, 2018.

\bibitem{qiu2021synface}
H.~Qiu, B.~Yu, D.~Gong, Z.~Li, W.~Liu, and D.~Tao.
\newblock Syn{F}ace: {F}ace recognition with synthetic data.
\newblock In {\em Proceedings of the IEEE/CVF International Conference on Computer Vision}, pages 10880--10890, 2021.

\bibitem{radford2021learning}
A.~Radford, J.~W. Kim, C.~Hallacy, A.~Ramesh, G.~Goh, S.~Agarwal, G.~Sastry, A.~Askell, P.~Mishkin, J.~Clark, et~al.
\newblock Learning transferable visual models from natural language supervision.
\newblock In {\em International Conference on Machine Learning}, pages 8748--8763. PMLR, 2021.

\bibitem{rombach2022high}
R.~Rombach, A.~Blattmann, D.~Lorenz, P.~Esser, and B.~Ommer.
\newblock High-resolution image synthesis with latent diffusion models.
\newblock In {\em Proceedings of the IEEE/CVF Conference on Computer Vision and Pattern Recognition}, pages 10684--10695, 2022.

\bibitem{ronneberger2015u}
O.~Ronneberger, P.~Fischer, and T.~Brox.
\newblock U-{N}et: {C}onvolutional networks for biomedical image segmentation.
\newblock In {\em MICCAI 2015: 18th International Conference on Medical Image Computing and Computer-Assisted Intervention, Munich, Germany, October 5-9, 2015, Proceedings, Part III 18}, pages 234--241. Springer, 2015.

\bibitem{saharia2022photorealistic}
C.~Saharia, W.~Chan, S.~Saxena, L.~Li, J.~Whang, E.~L. Denton, K.~Ghasemipour, R.~Gontijo~Lopes, B.~Karagol~Ayan, T.~Salimans, et~al.
\newblock Photorealistic text-to-image diffusion models with deep language understanding.
\newblock {\em Advances in Neural Information Processing Systems}, 35:36479--36494, 2022.

\bibitem{sauer2022stylegan}
A.~Sauer, K.~Schwarz, and A.~Geiger.
\newblock Style{GAN}-{XL}: {S}caling {S}tyle{GAN} to large diverse datasets.
\newblock In {\em ACM SIGGRAPH 2022 conference proceedings}, pages 1--10, 2022.

\bibitem{schuhmann2022laion}
C.~Schuhmann, R.~Beaumont, R.~Vencu, C.~Gordon, R.~Wightman, M.~Cherti, T.~Coombes, A.~Katta, C.~Mullis, M.~Wortsman, et~al.
\newblock {LAION-5b: A}n open large-scale dataset for training next generation image-text models.
\newblock {\em Advances in Neural Information Processing Systems}, 35:25278--25294, 2022.

\bibitem{segaran2021does}
K.~Segaran, A.~Z. Mohamad~Ali, and T.~W. Hoe.
\newblock Does avatar design in educational games promote a positive emotional experience among learners?
\newblock {\em E-learning and Digital Media}, 18(5):422--440, 2021.

\bibitem{serengil2020lightface}
S.~I. Serengil and A.~Ozpinar.
\newblock Light{F}ace: {A} hybrid deep face recognition framework.
\newblock In {\em 2020 Innovations in Intelligent Systems and Applications Conference (ASYU)}, pages 1--5. IEEE, 2020.

\bibitem{shen2023overview}
X.~Shen, S.~Lei, and J.~Liu.
\newblock Overview of cartoon face generation.
\newblock In {\em 2023 IEEE 6th Information Technology, Networking, Electronic and Automation Control Conference (ITNEC)}, volume~6, pages 792--799. IEEE, 2023.

\bibitem{somepalli2022diffusion}
G.~Somepalli, V.~Singla, M.~Goldblum, J.~Geiping, and T.~Goldstein.
\newblock Diffusion art or digital forgery? {I}nvestigating data replication in diffusion models.
\newblock In {\em Proceedings of the IEEE/CVF Conference on Computer Vision and Pattern Recognition}, 2023.

\bibitem{somepalli2023understanding}
G.~Somepalli, V.~Singla, M.~Goldblum, J.~Geiping, and T.~Goldstein.
\newblock Understanding and mitigating copying in diffusion models.
\newblock {\em Advances in Neural Information Processing Systems}, 36:47783--47803, 2023.

\bibitem{277172}
T.~Stadler, B.~Oprisanu, and C.~Troncoso.
\newblock {Synthetic Data {\textendash} Anonymisation Groundhog Day}.
\newblock In {\em 31st USENIX Security Symposium (USENIX Security 22)}, pages 1451--1468, Boston, MA, Aug. 2022. USENIX Association.

\bibitem{tripathy2020icface}
S.~Tripathy, J.~Kannala, and E.~Rahtu.
\newblock {IC}face: {I}nterpretable and controllable face reenactment using {GAN}s.
\newblock In {\em Proceedings of the IEEE/CVF Winter Conference on Applications of Computer Vision}, pages 3385--3394, 2020.

\bibitem{Webster2021ThisP}
R.~Webster, J.~Rabin, L.~Simon, and F.~Jurie.
\newblock This person (probably) exists. {I}dentity membership attacks against {GAN} generated faces.
\newblock {\em ArXiv}, abs/2107.06018, 2021.

\bibitem{wood2021fake}
E.~Wood, T.~Baltru{\v{s}}aitis, C.~Hewitt, S.~Dziadzio, T.~J. Cashman, and J.~Shotton.
\newblock Fake it till you make it: {F}ace analysis in the wild using synthetic data alone.
\newblock In {\em Proceedings of the IEEE/CVF International Conference on Computer Vision}, pages 3681--3691, 2021.

\bibitem{zhang2023adding}
L.~Zhang, A.~Rao, and M.~Agrawala.
\newblock Adding conditional control to text-to-image diffusion models.
\newblock In {\em Proceedings of the IEEE/CVF International Conference on Computer Vision}, pages 3836--3847, 2023.

\bibitem{zhang2018perceptual}
R.~Zhang, P.~Isola, A.~A. Efros, E.~Shechtman, and O.~Wang.
\newblock The unreasonable effectiveness of deep features as a perceptual metric.
\newblock In {\em Proceedings of the IEEE/CVF International Conference on Computer Vision (CVPR)}, 2018.

\bibitem{zhao2021action}
Y.~Zhao, L.~Yang, E.~Pei, M.~C. Oveneke, M.~Alioscha-Perez, L.~Li, D.~Jiang, and H.~Sahli.
\newblock Action unit driven facial expression synthesis from a single image with patch attentive {GAN}.
\newblock In {\em Computer Graphics Forum}, volume~40, pages 47--61. Wiley Online Library, 2021.

\bibitem{zheng2023poster}
C.~Zheng, M.~Mendieta, and C.~Chen.
\newblock {POSTER: A} pyramid cross-fusion transformer network for facial expression recognition.
\newblock In {\em Proceedings of the IEEE/CVF International Conference on Computer Vision (CVPR)}, pages 3146--3155, 2023.

\bibitem{zheng2020cartoon}
Y.~Zheng, Y.~Zhao, M.~Ren, H.~Yan, X.~Lu, J.~Liu, and J.~Li.
\newblock Cartoon face recognition: {A} benchmark dataset.
\newblock In {\em Proceedings of the 28th ACM International Conference on Multimedia}, pages 2264--2272, 2020.

\end{thebibliography}

\end{document}